\documentclass[runningheads]{llncs}
\usepackage[T1]{fontenc}
\usepackage{graphicx,verbatim}
\usepackage{amsmath}
\usepackage{amssymb}
\usepackage{multirow}
\usepackage{booktabs}
\usepackage{subcaption}
\usepackage{marvosym}
\begin{document}
\title{PIME: Prototype-based Interpretable MCTS-Enhanced Brain Network Analysis for Disorder Diagnosis}
\titlerunning{PIEM: Prototype-based Interpretable MCTS-Enhanced Framework}
%

\author{Kunyu Zhang\inst{1} \and 
Yanwu Yang\inst{2}\textsuperscript{\Letter} \and 
Jing Zhang\inst{3} \and 
Xiangjie Shi\inst{4} \and 
Shujian Yu\inst{5,6}}
\authorrunning{Zhang et al.}
\institute{
Zhengzhou University, Zhengzhou, China \and
University Hospital T\"ubingen, T\"ubingen, Germany \\
\email{yangyanwu1111@gmail.com} \and
China University of Geosciences, Wuhan, China\and
University of Science and Technology Beijing, Beijing, China\and
Vrije Universiteit Amsterdam, Amsterdam, The Netherlands \and
UiT - The Arctic University of Norway, Troms\o, Norway
}
  
\maketitle              
\begin{abstract}
    Recent deep learning methods for fMRI-based diagnosis have achieved promising accuracy by modeling functional connectivity networks. However, standard approaches often struggle with noisy interactions, and conventional post-hoc attribution methods may lack reliability, potentially highlighting dataset-specific artifacts. To address these challenges, we introduce PIME, an interpretable framework that bridges intrinsic interpretability with minimal-sufficient subgraph optimization by integrating prototype-based classification and consistency training with structural perturbations during learning. This encourages a structured latent space and enables Monte Carlo Tree Search (MCTS) under a prototype-consistent objective to extract compact minimal-sufficient explanatory subgraphs post-training. Experiments on three benchmark fMRI datasets demonstrate that PIME achieves state-of-the-art performance. Furthermore, by constraining the search space via learned prototypes, PIME identifies critical brain regions that are consistent with established neuroimaging findings. Stability analysis shows 90\% reproducibility and consistent explanations across atlases.

\keywords{Explainable AI \and fMRI \and Brain disorder diagnosis \and Information bottleneck \and Prototype learning \and Graph neural networks}

\end{abstract}

\section{Introduction}

Brain disorders including autism spectrum disorder (ASD), Alzheimer's disease (AD), and attention deficit hyperactivity disorder (ADHD) exhibit complex neural signatures, making precise neurobiological characterization challenging~\cite{jiang2023neuroimaging,yu2023cortical,zhang2025mvho}. Resting-state functional magnetic resonance imaging (rs-fMRI) provides an in vivo, noninvasive window into brain function and has become a cornerstone of computational diagnostic frameworks. With the advent of deep learning, researchers have developed increasingly sophisticated models to analyze brain networks~\cite{yang2023mapping,qiu2024towards}. Early convolutional neural networks captured localized functional connectivity patterns, graph neural networks (GNNs)~\cite{kipf2016semi,velickovic2017graph,xu2018gin} advanced whole-brain relational modeling, and recent large-scale self-supervised methods~\cite{yang2024brainmass} further improved diagnostic performance.

However, existing deep learning approaches face two critical limitations: processing whole-brain networks without modeling regional redundancy can be computationally inefficient and may overfit spurious connections, potentially degrading generalization and prediction stability; moreover, limited interpretability leaves clinicians unable to verify whether learned patterns reflect neurobiological mechanisms. The Information Bottleneck (IB) framework~\cite{tishby2015deep} addresses this by learning compressed representations that preserve task-relevant information and discard noise, as demonstrated by BrainIB~\cite{brainib} and BrainIB++~\cite{brainib++}. Prototype-based IB methods such as PGIB~\cite{seo2023interpretable} further improve interpretability through case-based reasoning.

Despite these advances, existing explanations are still often expressed as soft attributions or masks and do not explicitly answer a clinically relevant question: which minimal set of brain regions is sufficient to preserve a diagnostic decision? In addition, many explanations lack a stable semantic anchor across subjects, making them vulnerable to dataset-specific artifacts rather than neurobiologically validated biomarkers.

To address this gap, PIME (\textbf{P}rototype-based \textbf{I}nterpretable \textbf{M}CTS-\textbf{E}nhanced Framework) combines prototype-based representation learning with post-training minimal-sufficient subgraph search. PIME promotes robust, distributed representations through consistency regularization with stochastic structural perturbations, and uses prototype consistency as a semantic anchor for explanation. This enables MCTS to identify compact explanatory subgraphs (typically 10 to 16 regions) that preserve model decisions while improving explanation stability.

Our contributions can be summarized as follows:
\begin{itemize}
    \item We develop PIME, a unified framework that encourages robust representations via consistency training with structural perturbations, while integrating Information Bottleneck compression and prototype-based classification to improve noise sensitivity and interpretability.
    \item We introduce an MCTS-based post-training search guided by prototype similarity, operating on a structured latent space to identify compact explanatory subgraphs with neurobiologically plausible interpretations.
    \item Extensive evaluation on ABIDE, ADNI, and ADHD-200 demonstrates that PIME achieves state-of-the-art diagnostic accuracy (72.47\% on leave-one-site-out autism classification) while identifying 10-16 critical regions that are consistent with established neuroimaging findings.
\end{itemize}

\section{Related Work}

\subsection{MCTS for Interpretability}
Unlike gradient-based methods that approximate importance via soft masks~\cite{ying2019gnnexplainer}, search-based approaches treat interpretability as a combinatorial optimization problem to identify exact subgraphs. Monte Carlo Tree Search (MCTS) has shown remarkable success in reinforcement learning~\cite{silver2016mastering} and recently in explaining graph models for molecular properties~\cite{shan2021reinforcement,wang2026damr}. By iteratively expanding search trees, MCTS can be less sensitive to local optima than gradient-based optimization in some settings. In neuroimaging, however, the search space is exponentially larger due to dense connectivity. PIME introduces a prototype-guided MCTS strategy for fMRI, using prototype similarity as high-level guidance to prune the search space and extract compact explanatory biomarkers.

\section{Methodology}
\begin{figure}[t]
    \centering
    \includegraphics[width=0.94\linewidth]{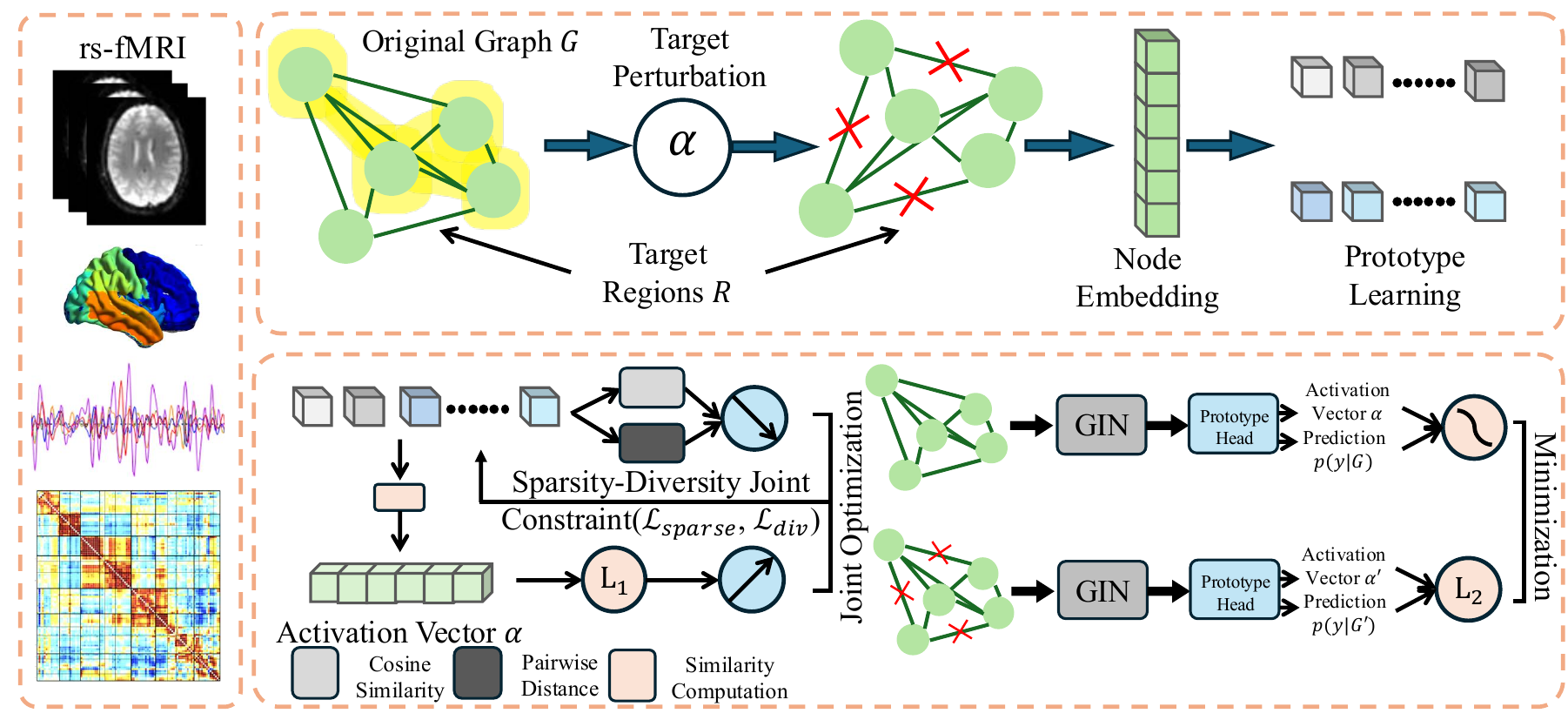}
    \caption{Overview of the PIME framework. It consists of a prototype-based IB training pipeline with consistency regularization and a post-training MCTS module for explanatory subgraph extraction.}
    \label{fig:framework}
\end{figure}

Figure~\ref{fig:framework} illustrates the architecture of PIME.
Consider a dataset of brain recordings $\{\mathbf{X}_i, y_i\}_{i=1}^N$, where each recording $\mathbf{X}_i \in \mathbb{R}^{C \times T}$ represents the blood-oxygen-level-dependent (BOLD) signal from resting-state fMRI for the $i$-th patient. Here, $C$ denotes the number of brain regions (e.g., 116 for AAL atlas~\cite{aal} or 200 for CC200 atlas~\cite{cc200}), $T$ indicates the temporal length, and $y_i \in \{1, \dots, K\}$ denotes the diagnostic label with $K$ classes. From each BOLD signal $\mathbf{X}_i$, we compute a functional connectivity matrix $A_i \in \mathbb{R}^{C \times C}$ where each entry $A_i^{jk}$ captures the Pearson correlation between brain regions $j$ and $k$. We then construct a brain network $G_i = (V, A_i)$ where $V$ represents $C$ brain regions as nodes and $A_i$ serves as the adjacency matrix. Our objective is to develop an interpretable classifier $f: G \rightarrow y$ that maps functional connectivity networks to diagnostic labels while identifying biologically meaningful biomarkers.

\subsection{Information Bottleneck Prototype Learning}
\label{sec:ib_proto}
High-dimensional noise in fMRI networks necessitates compressed representations that preserve task-relevant information. Given a functional connectivity network $G = (V, A)$ with $C$ ROIs, we use a variational surrogate of the Information Bottleneck objective. The GIN encoder predicts a Gaussian posterior $p_\theta(Z|G) = \mathcal{N}(\boldsymbol{\mu}_\theta(G), \boldsymbol{\sigma}_\theta(G))$, and the compression term is implemented via KL divergence to a Gaussian prior:
\begin{equation}
    \mathcal{L}_{IB} = \text{KL}(p_\theta(Z|G) \,\|\, r(Z))
\end{equation}

Unlike black-box neural networks, prototype-based classification enables interpretable predictions through case-based reasoning~\cite{zhang2025eva,wang2025protomol}: classifications rely on similarity to learnable exemplars representing typical disease patterns, allowing clinicians to understand decisions by examining which prototypes activate for each patient.

We define $M$ learnable prototypes $\mathcal{P} = \{\mathbf{p}_1, \dots, \mathbf{p}_M\}$ where $\mathbf{p}_k \in \mathbb{R}^d$ represents class-specific patterns~\cite{seo2023interpretable}. Each prototype $k$ is associated with learnable attention weights $\mathbf{a}_k \in \mathbb{R}^C$ that indicate the importance of each brain region for that prototype. A GIN encoder obtains graph embedding $\mathbf{z}$ via reparameterization during training.

The similarity between the embedding and each prototype is measured using the squared Euclidean distance:
\begin{equation}
    d_k(\mathbf{z}) = \|\mathbf{z} - \mathbf{p}_k\|_2^2
\end{equation}
The prototype activation is then computed via a softmax over negative distances:
\begin{equation}
    s_k(\mathbf{z}) = \frac{\exp(-d_k(\mathbf{z}))}{\sum_{j=1}^M \exp(-d_j(\mathbf{z}))}
\end{equation}
Let $\mathbf{s}(\mathbf{z}) = [s_1(\mathbf{z}), \dots, s_M(\mathbf{z})]^\top$. The predictive distribution is
\begin{equation}
    P(y|G) = \text{softmax}(W\,\mathbf{s}(\mathbf{z}) + b).
\end{equation}
The cross-entropy loss is
\begin{equation}
    \mathcal{L}_{ce} = -\frac{1}{N}\sum_{i=1}^N \log P(y_i|G_i).
\end{equation}

The prediction loss combines Cross-Entropy with prototype clustering and separation:
\begin{equation}
    \mathcal{L}_{pred}
    = \mathcal{L}_{ce}
    + \lambda_1 \frac{1}{N}\sum_{i=1}^N \min_{k \in \mathcal{P}_{y_i}} d_k(\mathbf{z}_i)
    + \lambda_2 \frac{1}{N}\sum_{i=1}^N \max\left(0, \Delta - \min_{k \notin \mathcal{P}_{y_i}} d_k(\mathbf{z}_i)\right).
\end{equation}
The second and third terms encourage samples to stay close to their class prototypes while remaining separated from other-class prototypes.
The base objective is $\mathcal{L}_{base} = \mathcal{L}_{pred} + \beta \mathcal{L}_{IB}$.

\subsection{Consistency Training and Regularization}
\label{sec:attack}
To reduce sensitivity to localized artifacts and encourage perturbation-invariant representations, we adopt consistency regularization with stochastic structural perturbations. For each graph $G_i$, we uniformly sample without replacement a node subset $\mathcal{R}_{rand}^{(i)} \subset \{1, \dots, C\}$ with $|\mathcal{R}_{rand}^{(i)}| = \lfloor r \cdot C \rfloor$. The binary mask $\mathbf{M}_{rand}^{(i)} \in \{0,1\}^{C \times C}$ is defined by $(\mathbf{M}_{rand}^{(i)})_{uv} = 1$ iff $u \in \mathcal{R}_{rand}^{(i)}$ or $v \in \mathcal{R}_{rand}^{(i)}$. The perturbed adjacency matrix is generated as $A_{i,pert} = A_i \odot (\mathbf{1} - \mathbf{M}_{rand}^{(i)})$, and the corresponding perturbed graph is $G_{i,pert} = (V, A_{i,pert})$.

To balance diagnostic accuracy with interpretability and stability under perturbations, we optimize:
\begin{equation}
    \min_{\theta} \quad \mathcal{L}_{total} = \mathcal{L}_{base} + \mathcal{R}(\theta)
\end{equation}
The regularization term $\mathcal{R}(\theta)$ combines structural consistency and geometric constraints:
\begin{equation}
\label{eq:reg}
\begin{aligned}
    \mathcal{R}(\theta) = \ & \lambda_{cons} \left( \text{KL}(P(y|G) \,\|\, P(y|G_{pert})) + \|\mathbf{s}(G) - \mathbf{s}(G_{pert})\|_2^2 \right) \\
    & + \lambda_{sparse} \frac{1}{M}\sum_{k=1}^M \|\mathbf{a}_k\|_1 - \lambda_{div} \frac{1}{M(M-1)}\sum_{i \neq j} \|\mathbf{p}_i - \mathbf{p}_j\|_2^2
\end{aligned}
\end{equation}
where $\lambda_{cons}$, $\lambda_{sparse}$, and $\lambda_{div}$ control the strength of consistency, attention sparsity, and prototype diversity, respectively. With a slight abuse of notation, $\mathbf{s}(G)$ denotes $\mathbf{s}(\mathbf{z}(G))$. Both $P(y|G_{pert})$ and $\mathbf{s}(G_{pert})$ are computed by forwarding $G_{i,pert}$ through the same encoder and prototype classification head used for $G_i$. The consistency term penalizes the discrepancy between the predictions and prototype activations of the original and perturbed graphs, encouraging invariance to structural perturbations rather than reliance on a small set of potentially brittle connections. The sparsity term applies $L_1$ regularization to the prototype attention weights $\mathbf{a}_k$ to encourage compact, human-interpretable region weighting. Finally, the diversity term increases pairwise distances between prototypes to reduce mode collapse and promote distinct prototype representations.

Post-training, we use MCTS to extract compact explanatory subgraphs by iteratively removing brain regions and scoring candidate subsets based on prototype similarity:
\begin{equation}
\label{eq:mcts_score}
    \text{score}(\mathcal{C}) = \log\left(\frac{d(\mathbf{z}_\mathcal{C}, \mathbf{p}^*) + 1}{d(\mathbf{z}_\mathcal{C}, \mathbf{p}^*) + \epsilon}\right)
\end{equation}
where $\mathcal{C}$ denotes the retained regions, $\mathbf{z}_\mathcal{C}$ is the masked graph embedding, and $\mathbf{p}^*$ is the predicted prototype. Since $\text{score}(\mathcal{C})$ is monotonically decreasing in $d(\mathbf{z}_\mathcal{C}, \mathbf{p}^*)$ (for $0<\epsilon<1$), maximizing it favors subsets whose embeddings remain close to the predicted prototype. MCTS returns the selected regions in $\mathcal{C}^*$ as the explanatory set under this objective.

\section{Experiments and Results}
\begin{table}[t]
    \centering
    \caption{Comparison with baselines on fMRI-based diagnosis. All values represent classification accuracy (\%). All shows mean $\pm$ std over leave-one-site-out or 5-fold CV. ADNI and ADHD-200 use 5-fold CV. ADNI tasks: 3-class (NC/MCI/AD classification), NC vs AD, MCI vs AD, and NC vs MCI. Best in bold, second-best underlined.}
    \label{tab:comparison}
    \resizebox{\textwidth}{!}{
    \begin{tabular}{lcccccc}
    \hline
    \textbf{Method} & \textbf{ABIDE} & \multicolumn{4}{c}{\textbf{ADNI}} & \textbf{ADHD-200} \\
    \cline{3-6}
     & & \textbf{3-class} & \textbf{NC vs AD} & \textbf{MCI vs AD} & \textbf{NC vs MCI} & \\
    \hline
    GCN~\cite{kipf2016semi} & 63.48{$\pm$7.61} & 58.27{$\pm$5.43} & 72.94{$\pm$4.72} & 71.63{$\pm$4.86} & 66.13{$\pm$5.21} & 63.72{$\pm$1.84} \\
    GAT~\cite{velickovic2017graph} & 67.92{$\pm$6.96} & 59.63{$\pm$5.18} & 72.19{$\pm$4.59} & 71.95{$\pm$4.71} & 66.28{$\pm$5.03} & 67.59{$\pm$1.52} \\
    GIN~\cite{xu2018gin} & 67.84{$\pm$6.79} & 60.12{$\pm$5.07} & 73.01{$\pm$4.51} & 72.68{$\pm$4.63} & 68.33{$\pm$4.89} & 67.81{$\pm$1.61} \\
    \hline
    Graph Transformer~\cite{shi2020masked} & 69.31{$\pm$6.53} & 59.37{$\pm$5.26} & 72.74{$\pm$4.68} & 72.79{$\pm$4.75} & 68.17{$\pm$5.12} & 69.38{$\pm$1.31} \\
    BrainNetTF~\cite{kan2022bnt} & \underline{71.43}{$\pm$4.67} & \underline{65.18}{$\pm$4.47} & \underline{78.42}{$\pm$3.83} & \underline{77.56}{$\pm$4.01} & \underline{72.89}{$\pm$4.23} & \underline{71.28}{$\pm$1.21} \\
    \hline
    SIB~\cite{SIB} & 61.73{$\pm$8.12} & 61.85{$\pm$4.81} & 75.23{$\pm$4.23} & 74.61{$\pm$4.37} & 70.12{$\pm$4.65} & 61.84{$\pm$2.13} \\
    DIR-GNN~\cite{dirgnn} & 66.82{$\pm$6.18} & 62.17{$\pm$4.73} & 75.81{$\pm$4.15} & 74.36{$\pm$4.29} & 70.63{$\pm$4.58} & 66.71{$\pm$1.73} \\
    ProtGNN~\cite{protgnn} & 64.51{$\pm$6.71} & 63.16{$\pm$4.62} & 76.49{$\pm$4.08} & 76.19{$\pm$4.21} & 71.59{$\pm$4.47} & 64.68{$\pm$1.89} \\
    PGIB~\cite{seo2023interpretable} & 64.37{$\pm$6.74} & 61.71{$\pm$4.69} & 75.24{$\pm$4.11} & 75.19{$\pm$4.35} & 71.01{$\pm$4.51} & 64.82{$\pm$1.96} \\
    BrainIB~\cite{brainib} & 70.22{$\pm$5.87} & 64.72{$\pm$4.51} & 77.26{$\pm$3.97} & 76.70{$\pm$4.13} & 72.47{$\pm$4.35} & 70.28{$\pm$1.42} \\
    BrainIB++~\cite{brainib++} & 70.63{$\pm$5.68} & 64.29{$\pm$4.56} & 78.13{$\pm$3.89} & 77.19{$\pm$4.08} & 72.61{$\pm$4.29} & 70.73{$\pm$1.28} \\
    \hline
    \textbf{PIME (Ours)} & \textbf{72.47}{$\pm$\textbf{5.12}} & \textbf{66.73}{$\pm$\textbf{4.38}} & \textbf{78.96}{$\pm$\textbf{3.76}} & \textbf{78.03}{$\pm$\textbf{3.91}} & \textbf{73.02}{$\pm$\textbf{4.17}} & \textbf{72.16}{$\pm$\textbf{1.14}} \\
    \hline
    \end{tabular}}
\end{table}

\begin{table}[t]
        \centering
        \caption{Ablation study on key components. All values represent classification accuracy (\%). Each variant removes one component while keeping others.}
        \label{tab:ablation}
        \resizebox{\textwidth}{!}{
        \begin{tabular}{lccccccc}
        \hline
        \textbf{Dataset} & \textbf{w/o IB} & \textbf{w/o Proto} & \textbf{w/o Cons} & \textbf{w/o Sparse} & \textbf{w/o Div} & \textbf{Baseline} & \textbf{Full} \\
        \hline
        ABIDE            & 70.10{$\pm$5.28} & 61.72{$\pm$6.43} & 67.54{$\pm$5.67} & 68.86{$\pm$5.41} & 70.57{$\pm$5.19} & 63.98{$\pm$6.21} & \textbf{72.47}{$\pm$\textbf{5.12}} \\
        ADNI (3-class)   & 63.18{$\pm$4.52} & 57.39{$\pm$5.67} & 61.52{$\pm$4.89} & 62.87{$\pm$4.63} & 64.05{$\pm$4.47} & 59.21{$\pm$5.41} & \textbf{66.73}{$\pm$\textbf{4.38}} \\
        ADHD-200         & 69.83{$\pm$1.26} & 61.54{$\pm$1.89} & 67.28{$\pm$1.53} & 68.79{$\pm$1.37} & 70.42{$\pm$1.21} & 64.37{$\pm$1.76} & \textbf{72.16}{$\pm$\textbf{1.14}} \\
        \hline
        \end{tabular}}
\end{table}
\subsection{Datasets and Experimental Settings}
We evaluate on three public fMRI datasets. ABIDE~\cite{abide} contains 505 controls and 530 autism subjects. ADNI~\cite{adni} contains 94 controls, 100 MCI, and 90 Alzheimer's subjects. ADHD-200~\cite{adhd200} contains 364 controls and 218 ADHD subjects. Following standard preprocessing protocols~\cite{cai2025mm,zeng2025knowledge}, brain regions are parcellated using AAL templates for ABIDE and ADHD-200, and CC200 templates for ADNI. Functional connectivity networks are constructed by computing Pearson correlation coefficients between all pairs of ROI time series from BOLD signals. We retain only the top 30\% strongest connections to reduce noise and computational cost, forming a sparse adjacency matrix $A$.

All experiments use PyTorch on NVIDIA A100 GPUs with a 3-layer GIN encoder having hidden dimensions of 128, 256, and 512, and latent dimension $d=64$. Training employs Adam optimization with learning rate $10^{-3}$, decay rate 0.5 every 50 epochs, and weight decay 0.03 for 200 epochs with batch size 32. The number of prototypes is $M = 7C$ where $C$ denotes the number of classes, yielding $M=14$ for binary tasks and $M=21$ for three-class ADNI. During consistency training, we randomly perturb $r=25\%$ of brain regions. Hyperparameters are selected via grid search over $\lambda_1, \lambda_2, \lambda_{cons}, \lambda_{sparse}, \lambda_{div} \in \{0.001, 0.01, 0.1\}$ and $\beta \in \{10^{-3}, 10^{-2}, 10^{-1}\}$, yielding $\beta=0.001$, $\lambda_1=\lambda_2=0.1$, $\lambda_{cons}=0.1$, $\lambda_{sparse}=\lambda_{div}=0.01$, $\Delta=1.0$. MCTS uses 20 rollouts with score constant $\epsilon=10^{-6}$.
Competing models use their recommended hyperparameters.
\begin{figure}[t]
    \centering
    \includegraphics[width=0.94\textwidth]{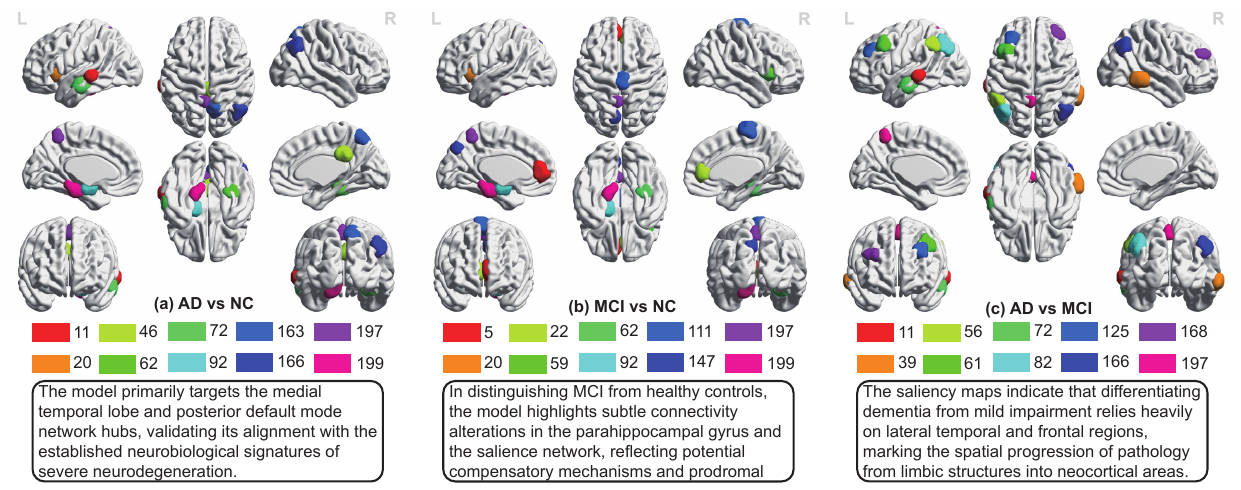}
    \caption{Top 10 MCTS-identified critical brain regions for each diagnostic task. (a) ADNI AD vs NC. (b) ADNI MCI vs NC. (c) ADNI AD vs MCI.}
    \label{fig:interpretability}
\end{figure}
\subsection{Experimental Results and Ablation Study}
We compare PIME with GNN backbones, Transformer baselines, and interpretable methods (see Table~\ref{tab:comparison}). All methods are evaluated using leave-one-site-out cross-validation on ABIDE and 5-fold cross-validation on ADNI and ADHD-200.

PIME consistently outperforms state-of-the-art methods across all benchmarks. On ABIDE and ADHD-200, our approach surpasses BrainIB++ by 1.84\% and 1.43\%, respectively. On ADNI, PIME achieves top performance in three-class staging (66.73\%) and early MCI detection (73.02\%) with superior stability.

Ablation studies systematically remove individual components to assess their contributions (Table~\ref{tab:ablation}). Prototype learning serves as the primary discriminative backbone, capturing the most critical class-specific patterns. The framework naturally fuses this with consistency training and Information Bottleneck regularization to enforce distributed biomarkers and filter high-dimensional noise, while diversity and sparsity constraints further refine feature distinctiveness, thereby consistently enhancing classification performance across all benchmarks.

\subsection{MCTS-Based Interpretability Analysis}

We apply MCTS post-training to extract compact explanatory subgraphs by iteratively pruning brain regions while monitoring prototype-guided prediction quality (Section~\ref{sec:attack}). In practice, MCTS typically selects about 10 regions out of 116 and about 16 regions out of 200 ROIs. Importantly, this compactness reflects an operational minimal-evidence explanation under our scoring objective: it identifies a small subset of regions sufficient to preserve the model's prototype-based decision, rather than implying that neurobiological mechanisms are intrinsically sparse or non-distributed.

Figure~\ref{fig:interpretability} visualizes the top-10 regions returned by MCTS for three binary classification tasks on ADNI. Starting from the full network, MCTS iteratively removes regions and evaluates the remaining subgraph via prototype-based prediction quality, returning a compact set of regions that best preserves the model's decision under the objective in Eq.~\ref{eq:mcts_score}.

\subsection{Prototype Activation Patterns and Stability Analysis}


We analyze top-5 activated regions for ADNI prototypes (MCI vs. NC) by mapping ROIs to dominant anatomy (Table~\ref{tab:prototype_regions}): NC prototypes emphasize posterior DMN and sensorimotor hubs, while MCI prototypes emphasize medial temporal and salience regions. Stability tests across prototype settings ($M=14$ vs. $28$) show 90\% Jaccard similarity on ADNI and ADHD-200 (Table~\ref{tab:stability_analysis}), and Table~\ref{tab:stability_atlas} shows cross-atlas stability on the same ADNI subset (MCI, $n=38$; NC, $n=37$); for CC200, each ROI is mapped to the anatomical region with the largest weight contribution.

\begin{table}[t]
\centering
\caption{Top-5 activated regions per prototype (ADNI). Representative prototypes: P1-3 (NC), P8-10 (MCI).}
\label{tab:prototype_regions}
\resizebox{\textwidth}{!}{
\begin{tabular}{c|ccccc}
\hline
\textbf{Prototype} & \textbf{Rank 1} & \textbf{Rank 2} & \textbf{Rank 3} & \textbf{Rank 4} & \textbf{Rank 5} \\
\hline
P1 (NC) & DMN: Precuneus L & DMN: Precuneus R & Visual: Calcarine R & Visual: Lingual L & DMN: Cuneus L \\
P2 (NC) & Motor: SMA R & Motor: Precentral L & Motor: Postcentral R & Auditory: Heschl L & Control: Parietal Inf \\
P3 (NC) & Visual: Occipital Mid & Visual: Fusiform L & DMN: PCC & Control: Frontal Mid & Visual: Calcarine L \\
P8 (MCI) & Limbic: ParaHippo R & Limbic: Hippo L & Limbic: ParaHippo L & Visual: Fusiform R & Temporal: Pole Mid \\
P9 (MCI) & SN: Insula L & SN: Insula R & SN: ACC L & SN: ACC R & Control: Frontal Inf \\
P10 (MCI) & SN: ACC R & Limbic: Amygdala L & SN: Frontal Med Orb & Subcortical: Thalamus & SN: Insula L \\
\hline
\end{tabular}}
\end{table}

\begin{table}[t]
    \centering
    \caption{Stability analysis.}
    \begin{subtable}[t]{0.57\textwidth}
        \centering
        \caption{Reproducibility of MCTS-identified functional regions across different prototype numbers}
        \label{tab:stability_analysis}
        \resizebox{\linewidth}{!}{
        \begin{tabular}{lcc}
        \toprule
        \textbf{Category} & \textbf{ADNI ROI IDs} & \textbf{ADHD ROI IDs} \\
        \midrule
        \textbf{Shared Regions} & \begin{tabular}[c]{@{}c@{}}5, 20, 22, 59, 62, \\ 92, 147, 197, 199\end{tabular} & \begin{tabular}[c]{@{}c@{}}29, 47, 49, 50, 51, \\ 79, 80, 83, 92\end{tabular} \\
        \midrule
        \textbf{M=14 Only} & 111 & 86 \\
        \textbf{M=28 Only} & 121 & 85 \\
        \midrule
        \textbf{Jaccard Similarity} & 90\% & 90\% \\
        \bottomrule
        \end{tabular}}
    \end{subtable}
    \begin{subtable}[t]{0.32\textwidth}
        \centering
        \caption{Cross-atlas stability on the same ADNI subjects}
        \label{tab:stability_atlas}
        \resizebox{\linewidth}{!}{
        \begin{tabular}{lc}
        \toprule
        \textbf{Method} & \textbf{Similarity} \\
        \midrule
        BrainIB & 30\% \\
        BrainIB++ & 60\% \\
        \textbf{PIME} & \textbf{80\%} \\
        \bottomrule
        \end{tabular}}
    \end{subtable}
\end{table}

\section{Conclusion and Future Work}


In summary, PIME unifies prototype-based IB learning with MCTS to bridge accurate fMRI diagnosis and reliable, minimal-sufficient explanations. Across three benchmarks, it achieves strong performance while extracting compact biomarkers (10–16 regions) with high reproducibility and cross-atlas consistency. Future work will develop multimodal prototype spaces and validate clinical robustness in prospective, multi-center settings.
%
%
%
%

\end{document}